# Constructive Incremental Learning for Fault Diagnosis of Rolling Bearings with Ensemble Domain Adaptation

Jiang Liu, Wei Dai, Senior *Member, IEEE*

*Abstract*—Given the prevalence of rolling bearing fault diagnosis as a practical issue across various working conditions, the limited availability of samples compounds the challenge. Additionally, the complexity of the external environment and the structure of rolling bearings often manifests faults characterized by randomness and fuzziness, hindering the effective extraction of fault characteristics and restricting the accuracy of fault diagnosis. To overcome these problems, this paper presents a novel approach termed constructive Incremental learning-based ensemble domain adaptation (CIL-EDA) approach. Specifically, it is implemented on stochastic configuration networks (SCN) to constructively improve its adaptive performance in multi-domains. Concretely, a cloud feature extraction method is employed in conjunction with wavelet packet decomposition (WPD) to capture the uncertainty of fault information from multiple resolution aspects. Subsequently, constructive Incremental learning-based domain adaptation (CIL-DA) is firstly developed to enhance the cross-domain learning capability of each hidden node through domain matching and construct a robust fault classifier by leveraging limited labeled data from both target and source domains. Finally, fault diagnosis results are obtained by a majority voting of CIL-EDA which integrates CIL-DA and parallel ensemble learning. Experimental results demonstrate that our CIL-DA outperforms several domain adaptation methods and CIL-EDA consistently outperforms state-of-art fault diagnosis methods in few-shot scenarios.

*Index Terms*—Domain adaptation, Incremental learning, bearing fault diagnosis, ensemble learning, limited data, domain matching

This work was supported in part by the Open Project Foundation of State Key Laboratory of Synthetical Automation for Process Industries under Grant 2020-KF-21-10, in part by the Natural Science Foundation of Jiangsu Province under Grant BK20200086, and in part by the National Natural Science Foundation of China under Grant 61973306. (Corresponding author: Wei Dai.)
Jiang Liu is with the School of Information and Control Engineering, China University of Mining and Technology, Xuzhou 221116, China (e-mail: liujiang2021@cumt.edu.cn).
Wei Dai is with the Artificial Intelligence Research Institute, China University of Mining Technology, Xuzhou 221116, China, and also with the State Key Laboratory of Synthetical Automation for Process Industries, Northeastern University, Shenyang 110819, China (e-mail: daiwei_neu@126.com).

## I. INTRODUCTION

Rolling bearings, a crucial part of rotating machinery in modern industry, not only support the machinery but also maintain the proper working position of shafts, thereby optimizing rotation accuracy during operation. Nevertheless, with the passage of time, rolling bearings become increasingly vulnerable to various types of failures induced by uncertain factors such as external temperature, dust, humidity, vibration, and so on. Its failures will result in a decrease of the dependability of machinery and even cause accidents and disasters. Therefore, both academia and industry place great emphasis on the research of bearing fault diagnosis [1].

The realm of intelligent bearing diagnosis has witnessed considerable attention towards deep learner models in recent times [2]. Notably, various deep networks, including convolutional neural network (CNN) [3]-[4], recurrent neural network (RNN) [5], deep belief network (DBN) [6], and auto encoder (AE) [7], have been deeply explored as excellent frameworks for fault diagnosis. These methods adeptly leverage end-to-end learning technology to autonomously extract highly discriminative fault-related features and establish a nonlinear mapping from input features to fault patterns. However, certain challenges persist, such as the time-consuming nature of training and the reliance on large amounts of data. Regrettably, rotating machinery equipment predominantly operates in a normal state, with instances of breakdown being infrequent. Consequently, it becomes arduous to accumulate an adequate number of fault samples. Thereby, bearing fault diagnosis represents a prototypical few-shot classification issue [8]. Consequently, deep learner models often encounter limitations in accuracy and practical applicability, struggling to capture effective feature representations and train a substantial number of model parameters in few-shot scenarios.

Combining specific feature extraction techniques for bearing vibration signals, shallow learner models, including support vector machine (SVM) [9], random vector functional link network (RVFLN) [10], back propagation neural networks (BPNN) [11], stochastic configuration network (SCN) [12], etc., offer viable alternatives for intelligent bearing diagnosis. SVM is commonly utilized for binary classification tasks, but the computational complexity of the classifier escalates with the number of categories when addressing multiple classification problems. The gradient descent-based training algorithms of BPNN often get stuck in local minima and incur slow convergence, and determining the appropriate number of network layers and nodes poses an open problem across different tasks. As a randomized learner model, RVFLN faces two crucial issues that hinder its practicality in fault diagnosis: the need for manual determination of hidden node numbers and the inefficient parameter assignment for hidden nodes. Conversely, SCNs [13], as a class of stochastic incremental networks, offer distinct advantages as their hidden nodes are determined randomly but constrainedly through a supervisory mechanism that ensures rapid convergence of prediction error with each incremental increase in the number of nodes. Furthermore, SCN sets a maximum number of configurable hidden nodes instead of a fixed number. Thus, SCNs showcase superiority in terms of random parameter selection, learning efficiency and generalization ability. However, the assumption of independent and identically distributed data, which underlies these shallow learner models, necessitates a consistent distribution of training and testing data. To our best knowledge, there is a lack of prior research on the domain adaptive learning of random incremental learner models, thereby limiting their broader application in cross-domain tasks.

Consequently, domain adaptation (DA), a special case of transfer learning (TL), has been introduced to fault diagnosis, exhibiting promising detection results even in few-shot scenarios [14]-[15]. The three main categories commonly used for domain adaptation are feature matching, instance reweighting, and classifier-based domain adaptation. The majority of feature matching-based methods solely focus on the data representation, followed by classifier selection, neglecting the potential benefits of integrating classifier design with the feature matching process into a unified paradigm [16]. Instance reweighting-based methods require equivalent conditional distributions between the source and target domains, as well as the reuse of several samples from the source domain in target domain learning [17]. The classifier-based methods [18]-[20] adapt the source classifier to the target by modifying model parameters while keeping data representation unchanged. Hence, they are well-suited for small domain shift problems and are beneficial to obtain a more compact model structure. Meanwhile, traditional incremental learning-based domain adaptation [21] is extensively studied in deep learning and is commonly employed to tackle the issue of forgetting existing knowledge in a data stream. However, this approach often requires a deeper network and poses challenges in achieving a compact framework.

For vibration signal analysis-based feature extraction, which is the mainstream method for fault diagnosis, numerous time-frequency domain methods, such as wavelet transform [22], empirical mode decomposition (EMD) [23], variational mode decomposition (VMD) [24], etc., have been investigated to effectively diagnose faults in rolling bearings. Despite contributing to the enhancement of fault diagnosis, these methods often encounter limitations imposed by the uncertainties present in feature information within practical applications. The uncertainties arise from the harsh working environment and complex mechanical structure of rotating machinery, as well as the random and fuzzy nature of faults, which make it so difficult to accurately represent fault information directly from collected signals that the higher accuracy of fault diagnosis is restricted. Distinct derivative entropies have been reported to address this problem with the development of nonlinear dynamic analysis theory, including appropriate entropy (ApEn) [25], sample entropy (SampEn) [26], and fuzzy entropy (FuzzyEn) [27]. These entropy-based nonlinear dynamic methods demonstrate excellent performance in analyzing vibration signals of rolling bearings. However, they face difficulties in parameter selection and exhibit other shortcomings, such as slight threshold variation causing sudden entropy changes that impact statistical stability.

To address the problems of fault diagnosis for rolling bearing in few-shot scenarios, this paper proposes constructive incremental learning-based ensemble domain adaptation (CIL-EDA). In CIL-EDA, cloud spatial features, including expectation, entropy and hyper entropy of cloud model (CM) [28], are extracted to express the intension of fault concept after multi-level wavelet packet decomposition (WPD). This approach captures both the randomness and fuzziness of bearing faults while comprehensively representing nonstationary signals without requiring hyperparameters in entropy calculation. Inspired by Adapt-SVM [18] and matching-based knowledge adaptation, a framework called constructive incremental learning-based domain adaptation (CIL-DA) is developed to advance the cross-domain learning capability of each hidden node in stochastic incremental networks, such as SCNs, and introduce a novel supervisory mechanism to ensure the convergence of CIL-DA. To robustly and accurately identify the health states of rolling bearings, ensemble learning is combined with CIL-DA using a limited number of labeled data from varying operating conditions. The main works of this paper are summarized as follows:

1) A novel CIL-DA framework based on the structural risk minimization principle and the regularization theory is first developed in constructive incremental learning. This framework incorporates a novel supervisory mechanism that endows each hidden node with cross-domain capability in terms of domain matching, and its corresponding convergence proof is provided.
2) To accurately represent the randomness and fuzziness of bearings fault information without much human intervention, cloud features are extracted from multi-level wavelet packet coefficients of small samples from each domain using a backward cloud generator of CM. This enables the acquisition of nonstationary fault-sensitive and nonlinear fault feature vectors.
3) To enhance the generalization performance of the fault classifier with limited data from the target domain, CIL-EDA is proposed to automatically and efficiently diagnose faults of rolling bearings. The base classifier of CIL-EDA takes as input a small number of labeled instances from multiple source domains acquired under varying operating conditions, along with limited labeled samples from the target domain.
4) Experiments on domain adaptation, conducted using rolling bearing data from various working conditions, confirm the effectiveness of CIL-DA in cross-domain fault diagnosis. Extensive experiments highlighted the outstanding diagnostic performance of CIL-EDA in few-shot scenarios, surpassing state-of-the-art fault diagnosis methods.

The rest of this paper is organized as follows. The relevant modeling method is briefly reviewed in Section II. In Section III, the proposed CIL-DA is illustrated in detail, including theoretical analysis and algorithm description. In Section IV, CIL-EDA is introduced for bearing fault diagnosis. Experiments are carried out to evaluate the proposed methods in Section V. Section VI draws our concluding remark, finally.

## II. RELATED WORK

Stochastic incremental networks, such as SCN and IRVFLN [19], attract a lot of attention because of their rapid modeling capabilities and impressive generalization performance. SCN is a classical constructive incremental learner model and will be elaborated upon in this section. Three types of SCN algorithms, including SC-I, SC-II and SC-III, have been proposed in [13]. The SCN-III outperforms to the other two algorithms in both learning speed and generalization ability, while SCN-I best embodies the innovative contribution of this article due to the first developed supervisory mechanism based on prediction error convergence. Thereby, SCN-I and SCN-III be overviewed.

Given a target function $f: R^d \to R^m$ and a dataset $D = (X, Y) = \{(X_i, Y_i), X_i \in R^d, Y_i \in R^m\}_{i=1}^N$, where $X$ represents the input data and the corresponding output data is denoted by $Y$. Assuming that a SCN with $L-1$ hidden nodes has been built, the output of target function is

$$f_{L-1}(X) = \sum_{j=1}^{L-1} \beta_j g_j(\omega_j, b_j, X) \quad (1)$$

where $g_j(\cdot)$ is the activation function of the $j$th hidden node, $\omega_j$ is the input weight for the $j$th hidden node, $b_j$ is the bias of the $j$th hidden node, and $\beta_j$ is the output weight between the $j$th hidden node and the output layer. $f_0$ is set to zero and denote the current residual error by $e_{L-1} = f - f_{L-1} = [e_{L-1,1}, \ldots, e_{L-1,m}]$, $L = 1, 2, \ldots$. If $\|e_{L-1}\|$ does not reach the tolerance error $e$, a new node $g_L(w_L, b_L)$ will be added to the original network, which leads its target function to $f_L = f_{L-1}(X) + \beta_L g_L(\omega_L, b_L, X)$. $h_L(X)$ represents the output vector of the $L$th hidden node in SCN, as presented in Equation (2).

$$h_L(X) = \left[g_L(\omega_L^T x_1 + b_L), \ldots, g_L(\omega_L^T x_N + b_L)\right]^T \quad (2)$$

Given $0 < r < 1$ and $0 < \mu_L < 1 - r$, the constraint of supervision mechanism that randomly assigns the hidden-node parameters $(w_L, b_L)$ is

$$\xi_L = \frac{\langle e_{L-1}^T(X) \cdot h_L(X) \rangle^2}{h_L^T(X) \cdot h_L(X)} - (1 - r - \mu_L) e_{L-1}^T(X) e_{L-1}(X) \geq 0. \quad (3)$$

For the SC-I algorithm, $\beta_L$ is calculated by

$$\beta_L = \frac{\langle e_{L-1}, g_L \rangle}{\|g_L\|^2} \quad (4)$$

while the global output weights $\beta = [\beta_1, \beta_2, \ldots, \beta_L]^T \in R^{L \times m}$ are evaluated by the least square method in the SC-III algorithm, as shown in Equation (5), where $H_L^\dagger$ is the Moore-Penrose generalized inverse and $\|\cdot\|_F$ represents the Frobenius norm.

$$\beta = \arg\min_{\beta} \frac{1}{2} \|Y - H_L^\dagger \beta\|_F^2 = H_L^\dagger Y \quad (5)$$

$H_L = [h_1, h_2, \ldots, h_L] \in R^{N \times L}$ is the output matrix of hidden layer of SCN with $L$ nodes. Finally, the output of SCN becomes

$$O_L = H_L \beta. \quad (6)$$

If the residual $\|e_L\|$ reaches the predefined tolerance threshold, the modeling process is deemed to be complete. Otherwise, a new node will continue to be added until the new residual meets the requirement. And it should be noted that SCN refers to SCN with SC-III algorithm in the remainder of this article. The proof of its convergence can be found in [13]. For more details on SCN, please refer to that paper.

## III. CONSTRUCTIVE INCREMENTAL LEARNING-BASED DOMAIN ADAPTATION

In this section, we comprehensively present the formulation of the proposed CIL-DA approach based on SCN, in which two algorithms, CILDA-I and CILDA-II, are introduced and their theoretical verifications of the convergence property are also given.

### A. CIL-DA

CIL-DA aims to learn a robust classifier of target domain in the manner of model incremental learning by utilizing all labeled samples which consist of a limited number of labeled samples from both the source and target domains. And its structural diagram is depicted in Fig. 1. Suppose that the abbreviations of source domain and target domain are represented by S and T, respectively. The framework of CIL-DA can be formulated as

$$\arg\min_{\beta, \xi_S^{j_S}, \xi_T^{j_T}} \frac{1}{2} \|\beta\|_2^2 + \frac{C_S}{2} \sum_{i=1}^{N_S} \|e_S^{j_S}\|_F^2 + \frac{C_T}{2} \sum_{j=1}^{N_T} \|e_T^{j_T}\|_F^2$$
$$+ \frac{\lambda}{2} D_{MMD}(X_S, X_T)^2 \quad (7)$$
$$s.t \begin{cases} H_S^{j_S} \beta = Y_S^{j_S} - e_S^{j_S}, j_S = 1, \ldots, N_S \\ H_T^{j_T} \beta = Y_T^{j_T} - e_T^{j_T}, j_T = 1, \ldots, N_T \end{cases}$$

where $H_S^{j_S} \in R^{1 \times L}$, $Y_S^{j_S}$ and $e_S^{j_S} \in R^{1 \times m}$ denote the output of hidden layer, the true label vector and the prediction error corresponding with the $j_S$th training sample from source domain; $H_T^{j_T} \in R^{1 \times L}$, $Y_T^{j_T}$ and $e_T^{j_T} \in R^{1 \times m}$ denote the output of hidden layer, the true label vector and the prediction error corresponding with the $j_T$th training sample from target domain; $N_S$ and $N_T$ represent the number of labeled training samples from source domain and target domain, respectively. And $C_S$, $C_T$ are the trade-off coefficients on the prediction errors of the labeled data from source and target domain, while $\lambda$ is the shrinkage regularization parameters on the Maximum Mean Discrepancy (MMD) [29].

The first three terms of Equation (7) tend to minimize the structural risk in terms of cross-domain. The divergence in distribution, particularly in conditional distribution, unfortunately remains significantly large. Thus, the standard classifier inferred by the first three terms may not generalize well to the target domain. Intuitively, it is reasonable to assume that the distributional characteristics $H_S$ and $H_T$ may become more similar following their mapping by the output weights $\beta$. Consequently, MMD is employed, which constitutes the final term of Equation (7), to mitigate the distributional divergence between the feature spaces of source and target domain mapped by SCN. As a distance metric method, $D_{MMD}(X_S, X_T)$ quantifies the dissimilarity between the empirical expectations of the mapped features from two domains, and it is defined as

$$D_{MMD} = \sum_{c=0}^{m} \left\| \left( \frac{1}{n_S^{(c)}} \sum_{x_{j_S} \in X_S^{(c)}} h_S\left(X_S^{j_S}\right) - \frac{1}{n_T^{(c)}} \sum_{x_{j_T} \in X_T^{(c)}} h_T\left(X_T^{j_T}\right) \right) \beta \right\|_F \quad (8)$$

where $n_S^{(0)} = N_S$, $n_T^{(0)} = N_T$; $n_S^{(c)}$ and $n_T^{(c)}$ indicate the number of labeled samples belonging to the $c$th class in the source domain and target domain when $c \neq 0$, respectively. $X_S^{(c)}$ and $X_T^{(c)}$ represent the overall samples from the $c$th class in the source domain and target domain, respectively. Noteworthily, Equation (8) demonstrates that the marginal distribution distance of source and target domain is minimized when $c = 0$, while simultaneously reducing the conditional probability distribution distance by bringing their intra-class centroids closer when $c \neq 0$. And Equation (8) can be further denoted by

$$D_{MMD} = \sum_{c=0}^{m} \left\| \left( D_S^{(c)} - D_T^{(c)} \right) \beta \right\|_F = \sum_{c=0}^{m} \left\| \Delta d^{(k)} \beta \right\|_F. \quad (9)$$

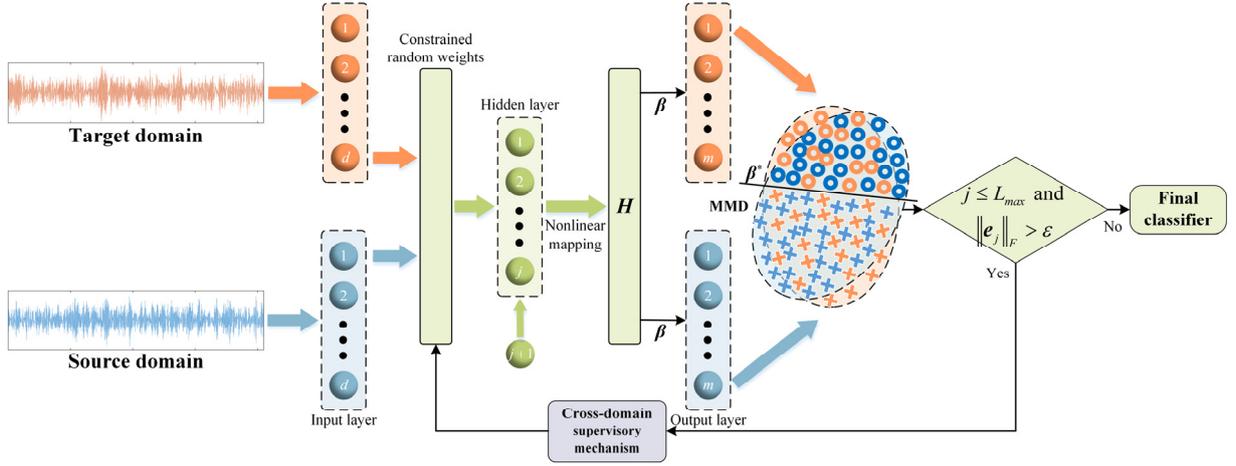

Fig. 1. Structural diagram of the proposed CIL-DA

Hence, $D^{(0)}$ signifies the center of the feature space that is mapped by the hidden layer of SCN from two distinct domains, while $D^{(c)}(c \neq 0)$ represents the center of the $c$th class in the mapped feature space.

In a nutshell, CIL-DA minimizes the structural risk functional and the joint adaptation of both marginal and conditional distributions to cope with the considerable shifts in data distributions across different domains in the manner of constructive incremental learning, ultimately learns an adaptive classifier for multi-class problems.

### B. CILDA-$\mathcal{I}$ Algorithm

Suppose that a CIL-DA containing $L$-1 hidden nodes has been built, its output weights of current network are

$$\beta = [\beta_1, \beta_2, \ldots, \beta_{L-1}]^T \quad (10)$$

where $\beta \in R^{(L-1) \times m}$. Then, an adaptive scheme needs to be developed to assign appropriate parameters $(w_L, b_L)$ to the $L$th hidden node and calculate the output weight $\beta_L$ after adding a new hidden node, which will result in higher model prediction accuracy for the CIL-DA with $L$ hidden nodes. Equation (8) can be transformed into the objective cost function, denoted as

$$f(\beta_L) = \frac{1}{2}[\beta \quad \beta_L]\begin{bmatrix}\beta \\ \beta_L\end{bmatrix} + \frac{C_S}{2}\|e_{S,L}\|^2 + \frac{C_T}{2}\|e_{T,L}\|^2$$
$$+ \frac{\lambda}{2}\left\|[d \quad d_L]\begin{bmatrix}\beta \\ \beta_L\end{bmatrix}\right\|^2$$
$$= \frac{1}{2}\|\beta\|_2^2 + \frac{1}{2}\|\beta_L\|_2^2 + \frac{C_S}{2}\|e_{S,L}\|_F^2 + \frac{C_T}{2}\|e_{T,L}\|_F^2 + \frac{\lambda}{2}\|d\beta + d_L\beta_L\|_F^2 \quad (11)$$

where $d \in R^{(m+1) \times (L-1)}$. $\beta_L$ can be solved by setting the derivative of the objective function with respect to $\beta_L$ to be zero by incorporating $e_{S,L} = e_{S,L-1} - g_{S,L}\beta_L$ and $e_{T,L} = e_{T,L-1} - g_{T,L}\beta_L$. Afterward, we obtain

$$\frac{\partial f(\beta_L)}{\partial \beta_L} = \beta_L - C_S g_{S,L}^T(e_{S,L-1} - g_{S,L}\beta_L)$$
$$- C_T g_{T,L}^T(e_{T,L-1} - g_{T,L}\beta_L) + \lambda d_L^T(d\beta + d_L\beta_L) = 0. \quad (12)$$

Thus, the optimal $\beta_L$ is written as

$$\beta_L = \frac{C_S g_{S,L}^T e_{S,L-1} + C_T g_{T,L}^T e_{T,L-1} - \lambda d_L^T d\beta}{1 + C_S \|g_{S,L}\|^2 + C_T \|g_{T,L}\|^2 + \lambda \|d_L\|^2} \quad (13)$$

and $\beta_{L,q}$ is further denoted as

$$\beta_{L,q} = \frac{\frac{C_S}{C_T}\langle e_{S,L-1,q}, g_{S,L}\rangle + \langle e_{T,L-1,q}, g_{T,L}\rangle - \frac{\lambda}{C_T}\langle d^T d_L, \beta_q\rangle}{\frac{1}{C_T} + \frac{C_S}{C_T}\|g_{S,L}\|^2 + \|g_{T,L}\|^2 + \frac{\lambda}{C_T}\|d_L\|^2} \quad (14)$$

where $q = 1,...,m$.

*Theorem*: Assuming that $span(\Gamma)$ is dense in $L_2$ space, $\forall g \in \Gamma$, $0 < \|g\| < \tau$ for some $\tau \in R^+$. Given three positive real numbers $(C_S, C_T, \lambda)$, $0 < r < 1$ and a non-negative real number sequence $\{\mu_L\}$ with $\lim_{L\to+\infty}\mu_L = 0$ and $\mu_L < 1 - r$. For $L = 1,2,...$ and $q = 1,2,...,m$, make the following definition indicated in Equation (15).

$$\delta_L = \sum_{q=1}^{m}\delta_{L,q}, \delta_{L,q} = (1 - r - \mu_L)\|e_{T,L-1,q}\|^2 \quad (15)$$

If random basis function $g_{S,L}$ and $g_{T,L}$ are generated to satisfy the following inequalities:

$$2G_L\langle e_{T,L-1,q}, g_{T,L}\rangle E_{L-1,q} - \|g_{T,L}\|^2 E_{L-1,q}^2 \geq b_g^2 \delta_{L,q} \quad (16)$$

where

$$E_{L-1,q} = \frac{C_S}{C_T}\langle e_{S,L-1,q}, g_{S,L}\rangle + \langle e_{T,L-1,q}, g_{T,L}\rangle - \frac{\lambda}{C_T}\langle d^T d_L, \beta_q\rangle,$$

$$G_L = \frac{1}{C_T} + \frac{C_S}{C_T}\langle g_{S,L}, g_{S,L}\rangle + \langle g_{T,L}, g_{T,L}\rangle + \frac{\lambda}{C_T}\langle d_L, d_L\rangle, \text{ and}$$

$$b_g = \frac{1}{C_T} + \left(\frac{C_S}{C_T} + 1 + \frac{\lambda}{C_T}\right)\tau^2,$$

the output weights are constructively evaluated by Equation (14). Finally, we have $\lim_{L\to+\infty}\|f - f_L\| = 0$.

*Proof*: Firstly, we have

$$\|e_{T,L}\|^2 - \|e_{T,L-1}\|^2 \quad L \to +\infty$$
$$= \sum_{q=1}^{m}\langle e_{T,L-1,q} - g_{T,L}\beta_{L,q}, e_{T,L-1,q} - g_{T,L}\beta_{L,q}\rangle$$
$$- \sum_{q=1}^{m}\langle e_{T,L-1,q}, e_{T,L-1,q}\rangle \quad (17)$$
$$= \sum_{q=1}^{m}[\langle g_{T,L}\beta_{L,q}, g_{T,L}\beta_{L,q}\rangle - 2\langle e_{T,L-1,q}, g_{T,L}\beta_{L,q}\rangle]$$

According to Equation (14), Equation (17) can be rewritten as

$$\|e_{T,L-1}\|^2 - \|e_{T,L}\|^2$$
$$= \sum_{q=1}^{m}\left(\frac{2}{C_T} + \frac{2C_S}{C_T}\|g_{S,L}\|^2 + \frac{2\lambda}{C_T}\|d_L\|^2 + \|g_{T,L}\|^2\right)\langle \beta_{L,q}, \beta_{L,q}\rangle$$
$$- 2\sum_{q=1}^{m}\left(\frac{B_g E_{L-1,q}}{B_g^2}\right)(E_{L-1,q} + \langle e_{T,L-1,q}, g_{T,L}\rangle) \quad (18)$$
$$= \sum_{q=1}^{m}\frac{2\langle e_{T,L-1,q}, g_{T,L}\rangle E_{L-1,q}}{B_g} - \|g_{T,L}\|^2\frac{E_{L-1,q}^2}{B_g^2}$$

where $B_g$ is calculated by

$$\frac{1}{C_T} + \frac{C_S}{C_T}\|g_{S,L}\|^2 + \|g_{T,L}\|^2 + \frac{\lambda}{C_T}\|d_L\|^2. \quad (19)$$

Recall that $2G_L\langle e_{T,L-1,q}, g_{T,L}\rangle E_{L-1,q} - \|g_{T,L}\|^2 E_{L-1,q}^2 \geq 0$ due to $\|e_{L-1,q}\|^2 \geq 0$ from Equation (16). Hence, Equation (18) is reformulated as

$$\|e_{T,L-1}\|^2 - \|e_{T,L}\|^2$$
$$= \sum_{q=1}^{m}\frac{2G_L\langle e_{T,L-1,q}, g_{T,L}\rangle E_{L-1,q} - \|g_{T,L}\|^2 E_{L-1,q}^2}{B_g^2} \quad (20)$$
$$\geq \sum_{q=1}^{m}\frac{2G_L\langle e_{T,L-1,q}, g_{T,L}\rangle E_{L-1,q} - \|g_{T,L}\|^2 E_{L-1,q}^2}{b_g^2} \geq 0$$

It is easy to verify that $\{\|e_{T,L}\|^2\}$ is monotonically decreasing according to Equation (20). Thus, the residual error $\|e_{T,L}\|$ of current CIL-DA is convergent as on target domain. From Equation (14)-(17), we can obtain

$$\|e_{T,L}\|^2 - (\gamma + \mu_L)\|e_{T,L-1}\|^2$$
$$= \sum_{q=1}^{m}\langle e_{T,L-1,q} - g_{T,L}\beta_{L,q}, e_{T,L-1,q} - g_{T,L}\beta_{L,q}\rangle$$
$$- (\gamma + \mu_L)\sum_{q=1}^{m}\langle e_{T,L-1,q}, e_{T,L-1,q}\rangle$$
$$= \sum_{q=1}^{m}(1 - \gamma - \mu_L)\langle e_{T,L-1,q}, e_{T,L-1,q}\rangle \quad (21)$$
$$+ \sum_{q=1}^{m}[\langle g_{T,L}\beta_{L,q}, g_{T,L}\beta_{L,q}\rangle - 2\langle e_{T,L-1,q}, g_{T,L}\beta_{L,q}\rangle]$$
$$= (1 - \gamma - \mu_L)\|e_{T,L-1}\|^2$$
$$- \sum_{q=1}^{m}\frac{2G_L\langle e_{T,L-1,q}, g_{T,L}\rangle E_{L-1,q} - \|g_{T,L}\|^2 E_{L-1,q}^2}{B_g^2}$$

According to Equation (20), Equation (21) is expressed as

$$\|e_{T,L}\|^2 - (\gamma + \mu_L)\|e_{T,L-1}\|^2$$
$$= \delta_L - \sum_{q=1}^{m}\frac{2G_L\langle e_{T,L-1,q}, g_{T,L}\rangle E_{L-1,q} - \|g_{T,L}\|^2 E_{L-1,q}^2}{B_g^2} \quad (22)$$
$$\leq \delta_L - \sum_{q=1}^{m}\frac{2G_L\langle e_{T,L-1,q}, g_{T,L}\rangle E_{L-1,q} - \|g_{T,L}\|^2 E_{L-1,q}^2}{b_g^2}$$
$$\leq 0$$

Therefore, the following inequality holds:

$$0 \leq \|e_{T,L}\|^2 \leq (\gamma + \mu_L)\|e_{T,L-1}\|^2 \leq \prod_{k=1}^{L-1}(\gamma + \mu_k)\|e_{T,0}\|^2. \quad (23)$$

Finally, $\lim_{L\to+\infty}\|e_{T,L}\|^2 = 0$ is obtained because of $\lim_{L\to+\infty}\prod_{k=1}^{L}(\gamma + \mu_k)\|e_{T,0}\|^2 = 0$, which implies $\lim_{L\to+\infty} e_{T,L} = 0$. The proof of Theorem of CILDA-I is now complete.

*Remark*: Theorem of CILDA-I provides us with a constructive incremental scheme for shallow networks in term of domain adaptation, primarily encompassing Equation (14) and (16), which consequently can bring a universal approximator in target domain. Equation (14) represents the cross-domain supervisory mechanism that effectively leverages knowledge from both source and target domains to search for the appropriate input weight $w_L$ and bias $b_L$ for a new hidden node of CIL-DA. And Equation (14) indicates that the random assignment of $w_L$ and $b_L$ are constrained through two distinct domains in the learning process of node increment. To the best of out knowledge, the cross-domain supervisory mechanism we propose is the first attempt in the design of incremental learning models, and it fills the gap in applying datasets of different distributions in SCNs

It is worth mentioning that $\boldsymbol{\beta}_L = [\beta_{L,1}, \beta_{L,1}, \ldots, \beta_{L,m}]^T$ in the Theorem of CILDA-I is analytically evaluated by Equation (14) and remains fixed for additional steps. However, this determination scheme may lead to a significantly slow convergence rate during the constructive process of CIL-DA. Consequently, inspired by original SCN, a recalculation scheme for the output weights is considered and will be detailed in the following subsection.

## C. CILDA-II Algorithm

Define $\boldsymbol{e}_{T,L}^* = f - \sum_{j=1}^{L} \boldsymbol{g}_{T,j} \boldsymbol{\beta}_j^*$ and $\boldsymbol{e}_{S,L}^* = f - \sum_{j=1}^{L} \boldsymbol{g}_{S,j} \boldsymbol{\beta}_j^*$, and denote the intermediate value of $\beta_{L,q}^*$ for $q = 1, 2, \ldots, m$ as

$$\widetilde{\beta}_{L,q} = \frac{\frac{C_S}{C_T}\langle \boldsymbol{e}_{S,L-1,q}^*, \boldsymbol{g}_{S,L}\rangle + \langle \boldsymbol{e}_{T,L-1,q}^*, \boldsymbol{g}_{T,L}\rangle - \frac{\lambda}{C_T}\langle d^T d_L, \boldsymbol{\beta}_q\rangle}{\frac{1}{C_T} + \frac{C_S}{C_T}\|\boldsymbol{g}_{S,L}\|^2 + \|\boldsymbol{g}_{T,L}\|^2 + \frac{\lambda}{C_T}\|d_L\|^2}. \quad (24)$$

In addition, define $\widetilde{\boldsymbol{e}}_{T,L} = \boldsymbol{e}_{T,L-1}^* - \boldsymbol{g}_{T,L}\widetilde{\boldsymbol{\beta}}_L$ where $\boldsymbol{e}_{T,0}^* = f$ and $\widetilde{\boldsymbol{\beta}}_L = [\widetilde{\beta}_{L,1}, \ldots, \widetilde{\beta}_{L,m}]$.

*Theorem*: Assuming that $span(\Gamma)$ is dense in $L_2$ space, $\forall \boldsymbol{g} \in \Gamma, 0 < \|\boldsymbol{g}\| < \tau$ for some $\tau \in R^+$. Given three positive real numbers $(C_S, C_T, \lambda), 0 < r < 1$ and a non-negative real number sequence $\{\mu_L\}$ with $\lim_{L \to +\infty} \mu_L = 0$ and $\mu_L < 1 - r$. For $L = 1, 2, \ldots$ and $q = 1, 2, \ldots, m$, make the following definition indicated in Equation (25).

$$\delta_L^* = \sum_{q=1}^{m} \delta_{L,q}^*, \delta_{L,q}^* = (1 - r - \mu_L)\|\boldsymbol{e}_{T,L-1,q}^*\|^2 \quad (25)$$

If random basis function $\boldsymbol{g}_{S,L}$ and $\boldsymbol{g}_{T,L}$ are generated to satisfy the following inequalities:

$$2G_L\langle \boldsymbol{e}_{T,L-1,q}, \boldsymbol{g}_{T,L}\rangle E_{L-1,q}^* - \|\boldsymbol{g}_{T,L}\|^2 E_{L-1,q}^{*2} \geq b_g^2 \delta_{L,q} \quad (26)$$

where

$E_{L-1,q}^* = \frac{C_S}{C_T}\langle \boldsymbol{e}_{S,L-1,q}^*, \boldsymbol{g}_{S,L}\rangle + \langle \boldsymbol{e}_{T,L-1,q}^*, \boldsymbol{g}_{T,L}\rangle - \frac{\lambda}{C_T}\langle d^T d_L, \boldsymbol{\beta}_q\rangle$,

$G_L = \frac{1}{C_T} + \frac{C_S}{C_T}\langle \boldsymbol{g}_{S,L}, \boldsymbol{g}_{S,L}\rangle + \langle \boldsymbol{g}_{T,L}, \boldsymbol{g}_{T,L}\rangle + \frac{\lambda}{C_T}\langle d_L, d_L\rangle$, and

$b_g = \frac{1}{C_T} + \left(\frac{C_S}{C_T} + 1 + \frac{\lambda}{C_T}\right)\tau^2$. Afterward, Equation (7) is written with unconstrained form as

$$\arg\min_{\boldsymbol{\beta}} f(\boldsymbol{\beta}) \quad (27)$$

while $f(\boldsymbol{\beta})$ is expressed as

$$\frac{1}{2}\|\boldsymbol{\beta}\|_2^2 + \frac{C_S}{2}\|H_S\boldsymbol{\beta} - Y_S\|_F^2 + \frac{C_T}{2}\|H_T\boldsymbol{\beta} - Y_T\|_F^2 + \frac{\lambda}{2}\|d\boldsymbol{\beta}\|^2. \quad (28)$$

And the output weight are evaluated by

$$\boldsymbol{\beta}^* = [\boldsymbol{\beta}_1^*, \boldsymbol{\beta}_2^*, \ldots, \boldsymbol{\beta}_L^*]^T = \arg\min_{\boldsymbol{\beta}} f(\boldsymbol{\beta}) \quad (29)$$

Finally, we have $\lim_{L \to +\infty} \|f - f_L^*\| = 0$.

*Proof:* It is easy to know that $\|\boldsymbol{e}_{T,L}^*\|^2 \leq \|\widetilde{\boldsymbol{e}}_{T,L}\|^2 = \|\boldsymbol{e}_{T,L-1}^* - \boldsymbol{g}_{T,L}\widetilde{\boldsymbol{\beta}}_L\|^2 \leq \|\widetilde{\boldsymbol{e}}_{T,L-1}\|^2$. Then, we have

$$\begin{aligned}
&\|\boldsymbol{e}_{T,L}^*\|^2 - \|\boldsymbol{e}_{T,L-1}^*\|^2 \leq \|\widetilde{\boldsymbol{e}}_{T,L}\|^2 - \|\boldsymbol{e}_{T,L-1}^*\|^2 \\
&= \sum_{q=1}^{m}\langle \boldsymbol{e}_{T,L-1,q}^* - \boldsymbol{g}_{T,L}\widetilde{\beta}_{L,q}, \boldsymbol{e}_{T,L-1,q}^* - \boldsymbol{g}_{T,L}\widetilde{\beta}_{L,q}\rangle \\
&\quad - \sum_{q=1}^{m}\langle \boldsymbol{e}_{T,L-1,q}^*, \boldsymbol{e}_{T,L-1,q}^*\rangle \\
&= \sum_{q=1}^{m}\left[\langle \boldsymbol{g}_{T,L}\widetilde{\beta}_{L,q}, \boldsymbol{g}_{T,L}\widetilde{\beta}_{L,q}\rangle - 2\langle \boldsymbol{e}_{T,L-1,q}^*, \boldsymbol{g}_{T,L}\widetilde{\beta}_{L,q}\rangle\right].
\end{aligned} \quad (30)$$

Recall that $2G_L\langle \boldsymbol{e}_{T,L-1,q}^*, \boldsymbol{g}_{T,L}\rangle E_{L-1,q}^* - \|\boldsymbol{g}_{T,L}\|^2 E_{L-1,q}^{*2} \geq 0$ due to $\|\boldsymbol{e}_{L-1,q}^*\|^2 \geq 0$ from Equation (26). According to Equation (24), Equation (30) is revised as

$$\begin{aligned}
&\|\boldsymbol{e}_{T,L-1}^*\|^2 - \|\boldsymbol{e}_{T,L}^*\|^2 \\
&= \sum_{q=1}^{m}\frac{2\langle \boldsymbol{e}_{T,L-1,q}^*, \boldsymbol{g}_{T,L}\rangle E_{L-1,q}^*}{B_g} - \|\boldsymbol{g}_{T,L}\|^2 \frac{E_{L-1,q}^{*2}}{B_g^2} \\
&\geq \sum_{q=1}^{m}\frac{2G_L\langle \boldsymbol{e}_{T,L-1,q}^*, \boldsymbol{g}_{T,L}\rangle E_{L-1,q}^* - \|\boldsymbol{g}_{T,L}\|^2 E_{L-1,q}^{*2}}{b_g^2} \geq 0
\end{aligned} \quad (31)$$

which indicates that $\{\|\boldsymbol{e}_{T,L}^*\|^2\}$ is monotonically decreasing and convergent as $L \to +\infty$. From Equation (24)-(26) and (31), we can obtain

$$\begin{aligned}
&\|\boldsymbol{e}_{T,L}^*\|^2 - (\gamma + \mu_L)\|\boldsymbol{e}_{T,L-1}^*\|^2 \leq \|\widetilde{\boldsymbol{e}}_{T,L}\|^2 - (\gamma + \mu_L)\|\boldsymbol{e}_{T,L-1}^*\|^2 \\
&= \sum_{q=1}^{m}\langle \boldsymbol{e}_{T,L-1,q}^* - \boldsymbol{g}_{T,L}\widetilde{\beta}_{L,q}, \boldsymbol{e}_{T,L-1,q}^* - \boldsymbol{g}_{T,L}\widetilde{\beta}_{L,q}\rangle \\
&\quad - (\gamma + \mu_L)\sum_{q=1}^{m}\langle \boldsymbol{e}_{T,L-1,q}^*, \boldsymbol{e}_{T,L-1,q}^*\rangle \\
&= \sum_{q=1}^{m}(1 - \gamma - \mu_L)\langle \boldsymbol{e}_{T,L-1,q}^*, \boldsymbol{e}_{T,L-1,q}^*\rangle \\
&\quad + \sum_{q=1}^{m}\left[\langle \boldsymbol{g}_{T,L}\widetilde{\beta}_{L,q}, \boldsymbol{g}_{T,L}\widetilde{\beta}_{L,q}\rangle - 2\langle \boldsymbol{e}_{T,L-1,q}^*, \boldsymbol{g}_{T,L}\widetilde{\beta}_{L,q}\rangle\right] \\
&= (1 - \gamma - \mu_L)\|\boldsymbol{e}_{T,L-1}^*\|^2 \\
&\quad + \sum_{q=1}^{m}\left[\langle \boldsymbol{g}_{T,L}\widetilde{\beta}_{L,q}, \boldsymbol{g}_{T,L}\widetilde{\beta}_{L,q}\rangle - 2\langle \boldsymbol{e}_{T,L-1,q}^*, \boldsymbol{g}_{T,L}\widetilde{\beta}_{L,q}\rangle\right] \\
&= (1 - \gamma - \mu_L)\|\boldsymbol{e}_{T,L-1}^*\|^2 \\
&\quad - \sum_{q=1}^{m}\frac{2G_L\langle \boldsymbol{e}_{T,L-1,q}^*, \boldsymbol{g}_{T,L}\rangle E_{L-1,q}^* - \|\boldsymbol{g}_{T,L}\|^2 E_{L-1,q}^{*2}}{B_g^2}.
\end{aligned} \quad (32)$$

Afterward, according to Equation (31), Equation (32) is further expressed as

$$\|e^*_{T,L}\|^2 - (\gamma + \mu_L)\|e^*_{T,L-1}\|^2 =$$

$$= \delta^*_L - \sum_{q=1}^{m} \frac{2G_L \langle e^*_{T,L-1,q}, g_{T,L} \rangle E^*_{L-1,q} - \|g_{T,L}\|^2 E^{*2}_{L-1,q}}{B_g^2}$$

$$\leq \delta^*_L - \sum_{q=1}^{m} \frac{2G_L \langle e^*_{T,L-1,q}, g_{T,L} \rangle E^*_{L-1,q} - \|g_{T,L}\|^2 E^{*2}_{L-1,q}}{b_g^2} \quad (33)$$

$$\leq 0.$$

Finally, $\lim_{L \to +\infty} e^*_{T,L} = 0$ is revealed using the similar arguments presented in the proof of Theorem CILDA-I, which completes the proof of Theorem CILDA-II.

*Remark:* the output weights $\beta^*$ of CILDA-II can be evaluated by setting the derivative of the target cost function $f(\beta)$ with respect to $\beta$ to be zero, as illustrated in Equation (34) and (35). It is evident that the output weights of CILDA-II undergo complete updates after the addition of a new node to CIL-DA and are solved through global optimization of the target function, which increases the decrease of network residual error $\|e_{T,L}\|$ and accelerates the convergence of CILDA-II algorithm. Moreover, CIL-DA improves prediction accuracy from the perspective of model-based incremental learning rather than data-based incremental learning [21]. Thus, our proposed method is more suitable for the few-shot classification issue.

$$\frac{\partial f(\beta)}{\beta} = \beta + C_S H_S^T (H_S \beta - Y_S) \quad (34)$$
$$+ C_T H_T^T (H_T \beta - Y_T) + \lambda d^T d\beta = 0$$

$$\beta^* = (I + C_S H_S^T H_S + C_T H_T^T H_T + \lambda d^T d)^{-1} \quad (35)$$
$$\times (C_S H_S^T Y_S + C_T H_T^T Y_T)$$

### D. Algorithm Description

The proposed CILDA-I and CILDA-II algorithms will be describe in this section. In general, the key components of our proposed CILDA algorithms are summarized as follows:
1) *Configuration of Hidden Parameters:* Randomly assigning the input weights and biases to satisfy either constraint (16) or (26), followed by adding the newly generated hidden node to the current network.
2) *Determination of Output Weights:* Evaluating the output weights of the current network based on (13) or (35).

Denote by $e_{T,L-1}(X_T) = [e_{T,L-1,1}(X_T), \ldots, e_{T,L-1,m}(X_T)] \in R^{N_T \times m}$ and $e_{S,L-1,1}(X_S) = [e_{S,L-1,1}(X_S), \ldots, e_{S,L-1,m}(X_S)] \in R^{N_S \times m}$ as the corresponding residual error vector before adding the $L$th hidden node in target domain and source domain, where $e_{T,L-1,q}(X_T) = [e_{T,L-1,q}(x_T^1), \ldots, e_{T,L-1,q}(x_T^{N_T})]^T \in R^{N_T}$ and $e_{S,L-1,q}(X_S) = [e_{S,L-1,q}(x_S^1), \ldots, e_{S,L-1,q}(x_S^{N_S})]^S \in R^{N_S}$. Define following formulas:

$$h_{S,L}(X) = [g_L(\omega_L^T X_S^1 + b_L), \ldots, g_L(\omega_L^T X_S^N + b_L)]^T, \quad (36)$$

$$h_{T,L}(X) = [g_L(\omega_L^T X_T^1 + b_L), \ldots, g_L(\omega_L^T X_T^N + b_L)]^T. \quad (37)$$

The current hidden layer output matrixes corresponding to the source domain and target domain can thus be express as $H_{S,L} = [h_{S,1}, h_{S,2}, \ldots, h_{S,L}]$ and $H_{T,L} = [h_{T,1}, h_{T,2}, \ldots, h_{T,L}]$. Then, (14) can be rewritten as

$$\beta_{L,q} = \frac{\frac{C_S}{C_T} e_{S,L-1}^T \cdot h_{S,L} + e_{T,L-1}^T \cdot h_{T,L} - \frac{\lambda}{C_T} d^T d \cdot \beta_{L-1,q}}{\frac{1}{C_T} + \frac{C_S}{C_T} h_{S,L}^T \cdot h_{S,L} + h_{T,L}^T \cdot h_{T,L} + \frac{\lambda}{C_T} d^T \cdot d_L} \quad (38)$$

where $\beta_{L-1}$ indicates the output weights before adding the $L$th hidden node. For the sake of simplicity, $\varsigma_{L,q}, q = 1, 2, \ldots, m$ is introduced and will be used in pseudo codes, as denoted in

$$\varsigma_{L,q} = \frac{2G_L(e_{T,L-1,q}^T \cdot h_{T,L})E_{L-1,q} - E_{L-1,q}(h_{T,L}^T \cdot h_{T,L})E_{L-1,q}}{G_L^2} \quad (39)$$
$$- (1 - \gamma - \mu_L)\|e_{T,L-1}^*\|^2$$

where $G_L = \frac{1}{C_T} + \frac{C_S}{C_T} h_{S,L}^T h_{S,L} + h_{T,L}^T h_{T,L} + \frac{\lambda}{C_T} d_L^T d_L$ and $E_{L-1,q} = \frac{C_S}{C_T} e_{S,L-1,q}^T h_{S,L} + e_{T,L-1,q}^T h_{T,L} - \frac{\lambda}{C_T} d_L^T d\beta_{L-1,q}$.

Recall that the suboptimal solution $\beta^*$ can be computed using Equation (27) in CILDA-II. Finally, the implementation procedure of CILDA-I and CILDA-II are summarized in **Algorithm 1** and **Algorithm 2**, respectively.

---

**Algorithm 1 CILDA-I**

Given training datasets $D_S = \{(X_S^{j_S}, Y_S^{j_S})\}_{j_S=1}^{N_S}, X_S^{j_S} \in R^d, j_S \in R^m$ and $D_T = \{(X_T^{j_T}, Y_T^{j_T})\}_{j_T=1}^{N_T}, X_T^{j_T} \in R^d, Y_T^{j_T} \in R^m$ in source domain and target domain, Set maximum number of hidden nodes $L_{max}$, expected error tolerance $\varepsilon$, maximum times of random configuration $T_{max}$ and regularization parameters $\{C_T, C_S, \lambda\}$; Choose a set of positive scalars $r = [\gamma_{min} : \Delta\gamma : \gamma_{max}]$;

1. Initialize $e_0 = [Y_T^1, Y_T^2, \ldots, Y_T^{N_T}]^T$, $0 < \gamma < 1$, $W = []$ and $\Omega = []$;
2. **While** $L \leq L_{max}$ AND $\|e_0\|_F > \varepsilon$, **Do**

   **Phase 1: Hidden Parameters Configuration (Step 3-20)**
3.     **For** $\theta \in r$, **Do**
4.         **For** $t = 1, 2, \ldots, T_{max}$, **Do**
5.             Randomly assign $w_L$ and $b_L$ from $[-\theta, \theta]^d$ and $[-\theta, \theta]$;
6.             Calculate $h_{S,L}$, $h_{T,L}$ and $\varsigma_{L,q}$ based on (36)-(37) and (39);
7.             Set $\mu_L = (1-\gamma)/(L+1)$;
8.             **If** $\min\{\varsigma_{L,1}, \varsigma_{L,2}, \ldots, \varsigma_{L,m}\} \geq 0$
9.               Save $w_L$ and $b_L$ in $W$, $\varsigma_L = \sum_{q=1}^{m} \varsigma_{L,q}$ in $\Omega$;
10.             **Else go back to Step 4**
11.             **End If**
12.         **End for** (corresponds to **Step 4**)
13.         **If** $W$ is not empty
14.             Find $w_L^*$ and $b_L^*$ that $w_L$ maximize $\varsigma_L$ in $\Omega$;
15.             Set $H_{S,L} = [h_{S,1}^*, \ldots, h_{S,L}^*]$ and $H_{T,L} = [h_{T,1}^*, \ldots, h_{T,L}^*]$
16.             **Break** (go to **Step 18**)
17.         **Else**

| 18. | randomly set $\tau \in (0, 1-\gamma)$, update $\gamma = \gamma + \tau$, return to step 4; |
| --- | --- |
| 19. | **End If** |
| 20. | **End for** (corresponds to **Step 3**) |
| | **Phase 2: Output Weights Determination** |
| 21. | Calculate $\beta_{L,q}, q = 1, 2, \ldots, m$ according to (38) and obtain $\beta_L = [\beta_{L,1}, \ldots, \beta_{L,m}]$; Calculate $e_L = e_{L-1} - h^*_{T,L} \beta_L$ |
| 22. | Renew $e_0 = e_L$; $L = L + 1$ |
| 23. | **End While** |
| 24. | **Return** $\beta_1, \ldots, \beta_L, w^* = [w^*_1, \ldots, w^*_L], b^* = [b^*_1, \ldots, b^*_L]$. |

| **Algorithm 2** CILDA-II |
| --- |
| Given the same items in **Algorithm 1**. |
| 1. Initialize $e_0 = [Y_T^1, Y_T^2, \ldots, Y_T^{N_T}]^T$, $0 < \gamma < 1$, $W = []$ and $\Omega = []$; |
| 2. **While** $L \leq L_{max}$ AND $\|e_0\|_F > \varepsilon$, **Do** |
| 3. Proceed **Phase 1 of Algorithm 1**; |
| 4. Obtain $H_{S,L} = [h^*_{S,1}, \ldots, h^*_{S,L}]$ and $H_{T,L} = [h^*_{T,1}, \ldots, h^*_{T,L}]$; |
| 5. Calculate $\beta^* = [\beta^*_1, \ldots, \beta^*_L]^T$ according to (35); |
| 6. Calculate $e_L = H_{T,L} \beta^* - Y_T$ |
| 7. Renew $e_0 = e_L$; $L = L + 1$ |
| 8. **End While** |
| 9. **Return** $\beta^*_1, \ldots, \beta^*_L, w^* = [w^*_1, \ldots, w^*_L], b^* = [b^*_1, \ldots, b^*_L]$. |

## IV. CIL-EDA FOR FAULT DIAGNOSIS OF ROLLING BEARINGS

### A. Overall Framework

The illustration and flowchart of the proposed method are depicted in Fig. 2, comprising three sequential stages: vibration signals acquisition, multilevel cloud feature extraction, and ensemble cross-domain learning for fault diagnosis. Vibration signals are measured using acceleration vibration sensors from our target rolling bearing under multiple operation conditions. To extract effective fault-sensitive features of rolling bearings, the raw vibration signals undergo decomposition into multilevel wavelet packet coefficients. Cloud features of each wavelet packet coefficient are obtained at different decomposition levels, resulting in initial vibration features that encompass abundant diagnosis information characterized by uncertainty and non-stationarity. Subsequently, the training dataset from the target domain is paired with multiple training datasets from source domains to form the input for the base classifier of CIL-EDA that aims to learn an ensemble cross-domain classifier by leveraging a small number of labeled samples from the target domain as well as available labeled samples from multiple source domains. It is worth noting that source domains indicate varying operation conditions of our target rolling bearing, except for the target operation condition. Eventually, the fault diagnosis result is predicted by CIL-EDA through a majority voting process, capitalizing on the robust cross-domain capability of CIL-DA.

### B. Multilevel Cloud Feature Extraction

In general, fault diagnosis based solely on the original signals is challenging due to the intricate and noisy nature of raw vibration signals in an industrial operating environment. Hence, feature extraction, including signal preprocessing, becomes crucial for fault diagnosis using shallow learner models. In this section, the combination of CM and WPD is employed to extract fault features.

WPD is a typical multiresolution analysis method that enables the decomposition of a signal into several sub-signals with identical bandwidths but different central frequencies [30]. It is superior for WPD to provide higher accuracy in the high-frequency components of signals and decompose them without redundant or missing information. WPD is particularly suitable for processing non-stationary vibration signals characterized by high frequencies and strong background noises in complex industrial environments and has been widely adopted in the field of fault diagnosis [31]-[32]. The process of wavelet packet decomposition is illustrated in Fig. 3 for a decomposition level of three. Wavelet packet coefficient in each node contains distinct and diverse fault information at different frequency bands while varying decomposition levels offer different fault-related information.

However, the uncertainty in signal acquisition, which results in randomness and fuzziness in fault information, is a common yet underappreciated phenomenon. This uncertainty can significantly impact the accuracy of feature extraction and diminish the precision of fault diagnosis. CM, comprising of a forward cloud generator (FCG) and a backward cloud generator (BCG), is a cognitive model that effectively addresses this issue by facilitating bidirectional transformation between the connotation and extension of qualitative concepts as depicted in Fig. 4. Regarding BCG, the intension $(Ex, En, He)$ of certain health state of rolling bearing can be derived from a certain amount of data, representing the process of acquiring knowledge from quantitative data. Cloud feature entropy $(En)$, a derivative entropy from CM, serves as a metric to gauge uncertainty and comprehensively capture dynamic fault feature information without requiring hyperparameters in the calculation [33]. Therefore, we extract multilevel cloud features from different decomposition levels of WPD to obtain comprehensive fault feature information in terms of multiresolution analysis.

On the whole, the procedures of multilevel cloud feature extraction are listed as follows:

1) *Signal preprocessing:* vibration signal of each health state of rolling bearing is denoised by the wavelet filter and then normalized by

$$\hat{x}_{p,q} = \frac{x_{p,q} - x_p^{min}}{x_p^{max} - x_p^{min}}, p = 1 \ldots m \quad (40)$$

where $m$ represents the total number of healthy states, $x_{p,q}$ is a vibration sensor measurement at the $q$th sampling point under the $p$th health state. $x_p^{min}$ and $x_p^{max}$ are the minimum and maximum values under the $p$th health state. Next, sample segmentation is performed using a sliding window.

2) *Wavelet packet decomposition:* Calculate the wavelet packet coefficients at different decomposition levels using WPD according to [31].

3) *Cloud feature extraction:* The BCG of the one-dimensional cloud model is applied to extract the fault features $(Ex, En, He)$ of rolling bearing from each sample. $x = [x_1, x_2, \cdots, x_n]$ is a sample of wavelet packet coefficient.

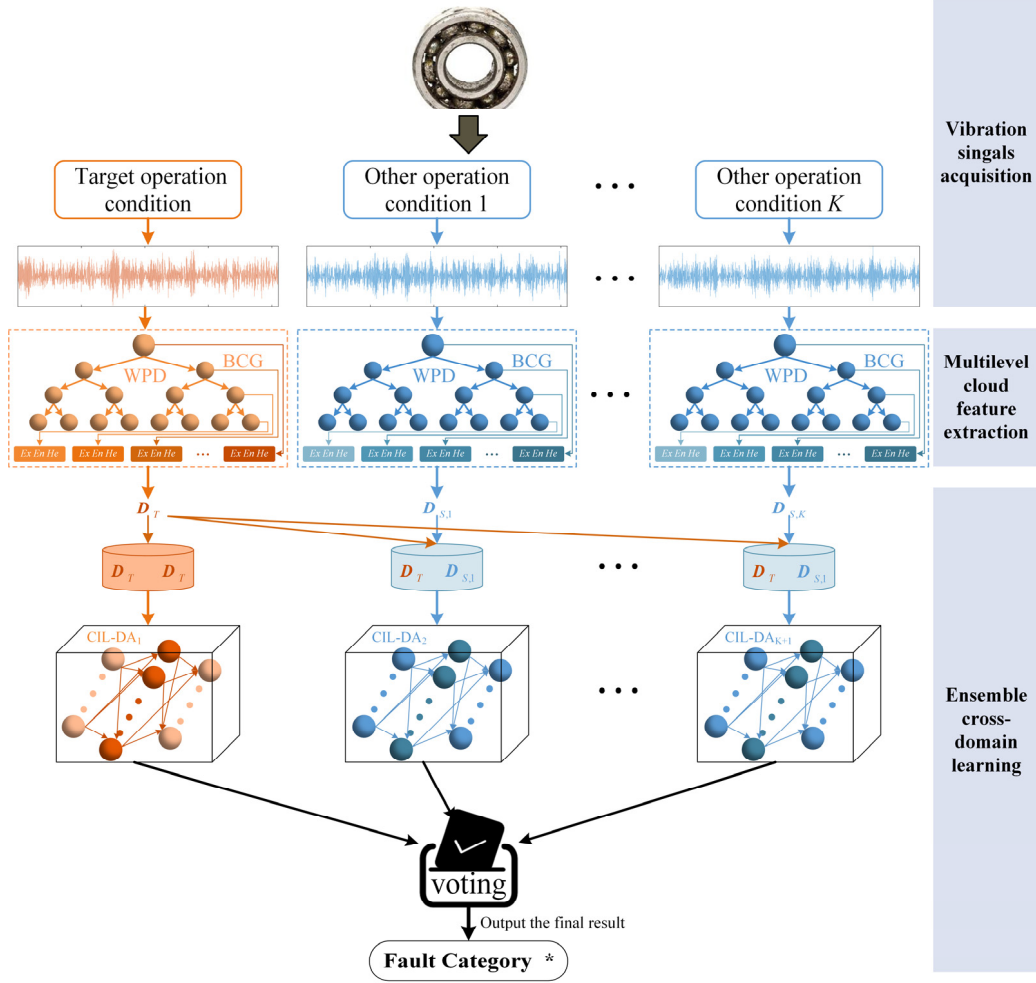

Fig. 2. Illustration and flowchart of the proposed CIL-EDA fault diagnosis method. $D_T$ denotes the dataset from target domain and $D_{S,k}$ ($k=1,\cdots,K$) represents the dataset from $k$th source domain.

Firstly, calculate its sample mean $\bar{X} = \frac{1}{n}\sum_{i=1}^{n} x_i$, first order center distance $\frac{1}{n}\sum_{i=1}^{n}|x_i - \bar{X}|$ and sample variance $S^2 = \frac{1}{n-1}\sum_{i=1}^{n}(x_i - \bar{X})^2$, Then, the expectation can be expressed as

$$Ex = \bar{X}. \quad (41)$$

And, calculate the feature entropy by

$$En = \sqrt{\frac{\pi}{2}} \times \frac{1}{n}\sum_{i=1}^{n}|x_i - Ex|. \quad (42)$$

Lastly, the hyper entropy is obtained by

$$He = \sqrt{|S^2 - En^2|}. \quad (43)$$

Thus, a feature set for one signal sample contains $\sum_{i=0}^{N} 3\cdot 2^i$ parameters under $N$-level decomposition.

### C. Ensemble Cross-domain Learning

In order to make full use of existing labeled data from multiple operation conditions of the target rolling bearing and simultaneously enhance the generalization performance of individual classifier, ensemble cross-domain learning founded on CIL-DA with parallel ensemble learning strategy is developed in this section. Vibration signals collected under the target operation condition and other operation conditions constitute the target domain $D_T$ and multiple source domains $D_{S,k}$ ($k=1,2\ldots,K$). A joint dataset obtained by combining $D_T$ with itself or $D_{S,k}$ in pairs acts as the input for base classifier of CIL-EDA, which also ensure the diversity among different base classifiers. Finally, the fault diagnosis result is predicted through a majority voting.

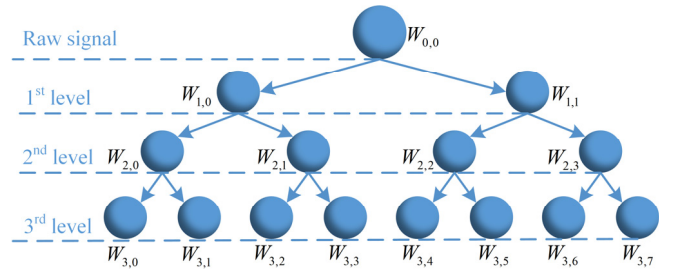

Fig. 3. Process of WPD when tdecomposition level is 3. $W_{i,j}$ represents wavelet packet coefficients at the $j$th subband of $i$th decomposition level.

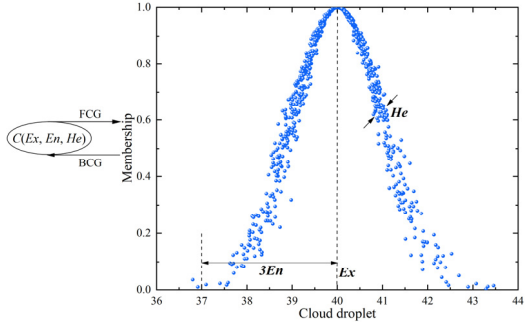

Fig. 4. Bidirectional cognitive transformation diagram of CM. $Ex$, $He$, and $He$ represent expected value, entropy and hyper entropy of CM, respectively.

TABLE I
DESCRIPTION OF SAMPLES UNDER VARYING WORKING CONDITIONS

| Dataset | Motor load | Health states | Number of training samples per class | Number of testing samples per class |
|---|---|---|---|---|
| A | 0 HP | REF | 5 | 30 |
|   |      | IRF | 5 | 30 |
|   |      | ORF | 5 | 30 |
|   |      | NS  | 25 | 30 |
| B | 1 HP | REF | 5 | 30 |
|   |      | IRF | 5 | 30 |
|   |      | ORF | 5 | 30 |
|   |      | NS  | 25 | 30 |
| C | 2 HP | REF | 5 | 30 |
|   |      | IRF | 5 | 30 |
|   |      | ORF | 5 | 30 |
|   |      | NS  | 25 | 30 |
| D | 3 HP | REF | 5 | 30 |
|   |      | IRF | 5 | 30 |
|   |      | ORF | 5 | 30 |
|   |      | NS  | 25 | 30 |

$f_k(X)$ $(k=1,2,\cdots,K+1)$ is the decision function of the $k$th CIL-DA in CIL-DA and $C_j (j=1,2,\cdots,m)$ is the label of the $j$th class. $Num_j = number\{k | f_k(x) = C_j\}$ means the total amount of $j$th label in all predictions from CIL-DA, where $number(\cdot)$ indicates counting the number. The final classification results can be obtained by

$$f_{final}(x) = \arg\max_j (Num_j) \quad (44)$$

## V. EXPERIMENTAL STUDY

In this section, the effectiveness and performance of the proposed approaches, including CIL-DA and CIL-EDA, are demonstrated using acceleration vibration signals of rolling bearings provided by the Bearing Data Center of Case Western Reserve University (CWRU) [34]. Original signals are acquired from the 6205-2RSJEM SKF deep groove ball bearing and the experimental platform consists of a 2 HP motor with drive end and fan end bearings, a torque sensor/encoder, and a dynamometer, as depicted in Fig. 5. Pitting corrosion is utilized to induce various bearing health states, including inner ring fault (IRF), outer ring fault (ORF) (with damage points positioned at 3 o'clock, 6 o'clock, and 12 o'clock), rolling element fault (REF), and normal state (NS), with fault diameters of 0.1778 mm (7 mils), 0.3556 mm (14 mils), and 0.5334 mm (21 mils). Furthermore, acceleration vibration signals are recorded under four distinct operation conditions corresponding to motor speeds of 1797 r/min with 0 HP load, 1772 r/min with 1 HP load, 1750 r/min with 2 HP load, and 1720 r/min with 3 HP load, employing a sampling frequency of 12kHz. All experiments are executed on the Windows 10 operating system (Intel i5-12600KF 3.69GHz CPU, 32GB RAM), and algorithms are coded in Python 3.9.

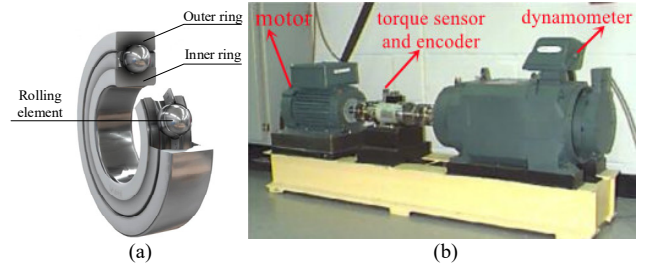

Fig. 5. Establishment of the experimental platform. (a) Deep groove ball bearing. (b) Experimental platform.

TABLE II
DESCRIPTION OF CLASSES FOR EACH DATASET

| Health state | Fault diameter (mils) | Labels |
|---|---|---|
| REF | 7  | 0 |
|     | 14 | 1 |
|     | 21 | 2 |
| IRF | 7  | 3 |
|     | 14 | 4 |
|     | 21 | 5 |
| ORF | 7  | 6 |
|     | 14 | 7 |
|     | 21 | 8 |
| NS  | -  | 9 |

### A. Effectiveness of CIL-DA

In this section, the performance of CIL-DA in domain adaptation will be showcased through experiments focused on fault diagnosis of rolling bearing under varying operation conditions. The raw signals obtained from the drive end rolling bearing are segmented by a sliding window with a length and step size of 1024. Subsequently, the segmented signals are processed utilizing our multilevel cloud feature extraction method to derive the feature datasets. Table I illustrates the division of the four feature datasets based on the four distinct operation conditions. Each training or testing sample is randomly selected from the original vibration signals. Within each operation condition, each fault state encompasses three fault diameter variations: 7 mils, 14 mils, and 21 mils, as displayed in Table II. Note that Table II complements Table I by providing information about the severity of each fault state. Moreover, training datasets consider the presence of imbalanced class distribution, which accurately reflects real-world scenarios.

The effectiveness and feasibility of the proposed method are validated by comparison with SCN, RVFL, SVM, and related domain adaptation approaches, namely UJD-RVFL [35], Adapt-SVM [18], CCSA [36], and CDAN [37]. It is evident that SCN, RVFL, and SVM, trained with limited labeled data from the target domain, are considered to prove the adaptability of CIL-DA and the aforementioned domain adaptation approaches.

TABLE III
DIAGNOSTIC PERFORMANCE OF ALL COMPARISON METHODS

| Cases | Methods | SCN | SVM | RVFL | CIL-DA | Adapt-SVM | UJD-RVFL | CCSA | CDAN |
|---|---|---|---|---|---|---|---|---|---|
| A→B | Accuracy (%) | 83.05 | 87.37 | 86.77 | **95.47** | 89.26 | 94.83 | 93.93 | 92.33 |
| | Time (s) | 1.46 | 0.0130 | 0.0068 | 11.49 | 0.15 | 24.33 | 35.16 | 42.07 |
| A→C | Accuracy (%) | 82.03 | 80.53 | 81.32 | 86.13 | 85.51 | 88.73 | 90.63 | **92.07** |
| | Time (s) | 1.14 | 0.0124 | 0.0077 | 11.32 | 0.15 | 26.33 | 35.63 | 44.22 |
| A→D | Accuracy (%) | 86.33 | 90.43 | 91.75 | **96.48** | 88.74 | 93.48 | 92.00 | 92.07 |
| | Time (s) | 1.66 | 0.0146 | 0.0082 | 12.17 | 0.13 | 25.66 | 35.09 | 44.00 |
| B→A | Accuracy (%) | 82.78 | 88.68 | 82.50 | **93.83** | 73.66 | 91.73 | 91.73 | 92.30 |
| | Time (s) | 1.75 | 0.0150 | 0.0084 | 16.47 | 0.13 | 26.03 | 35.53 | 45.06 |
| B→C | Accuracy (%) | 83.23 | 81.38 | 81.43 | **91.48** | 88.45 | 89.28 | 91.27 | 91.03 |
| | Time (s) | 1.28 | 0.0141 | 0.0086 | 11.52 | 0.12 | 26.58 | 36.71 | 45.40 |
| B→D | Accuracy (%) | 86.17 | 88.20 | 92.32 | **96.83** | 89.41 | 93.45 | 91.17 | 90.30 |
| | Time (s) | 1.44 | 0.0133 | 0.0091 | 12.42 | 0.13 | 27.73 | 36.09 | 44.46 |
| C→A | Accuracy (%) | 83.28 | 87.90 | 83.00 | **92.83** | 90.37 | 91.60 | 92.50 | 90.50 |
| | Time (s) | 1.72 | 0.0152 | 0.0083 | 12.06 | 0.11 | 23.16 | 37.07 | 46.11 |
| C→B | Accuracy (%) | 83.43 | 86.37 | 87.22 | **95.10** | 90.30 | 90.57 | 93.03 | 92.07 |
| | Time (s) | 1.61 | 0.0146 | 0.0085 | 12.53 | 0.13 | 23.58 | 37.21 | 46.66 |
| C→D | Accuracy (%) | 87.38 | 90.37 | 92.35 | **96.70** | 91.87 | 93.67 | 94.27 | 91.43 |
| | Time (s) | 1.48 | 1.0134 | 0.0107 | 13.08 | 0.14 | 24.32 | 38.68 | 45.18 |
| D→A | Accuracy (%) | 81.98 | 89.05 | 82.75 | **93.07** | 89.23 | 91.60 | 90.43 | 92.67 |
| | Time (s) | 1.80 | 0.0163 | 0.0098 | 12.68 | 0.12 | 25.71 | 39.35 | 48.98 |
| D→B | Accuracy (%) | 83.33 | 87.03 | 86.92 | **95.15** | 87.48 | 90.48 | 91.47 | 90.03 |
| | Time (s) | 1.59 | 0.0138 | 0.0105 | 12.30 | 0.15 | 24.16 | 39.14 | 48.93 |
| D→C | Accuracy (%) | 81.32 | 79.88 | 81.60 | 87.52 | 89.71 | 88.37 | **92.67** | 92.10 |
| | Time (s) | 1.28 | 0.0130 | 0.0102 | 13.44 | 0.14 | 26.11 | 40.55 | 48.74 |

Remarkably, all comparison approaches are employed in the context of supervised tasks, and a cross-domain diagnosis task can be regarded as $A \rightarrow B$, where A represents Dataset A in the source domain and B represents Dataset B in the target domain. For SCN and CIL-DA, their maximum times of random configuration are set as $T_{max} = 100$, expected error tolerance is set as $\varepsilon = 0.1$, random parameter range is set as $\gamma = [0.5, 1, 3, 5, 7, 10, 25, 50, 100, 150, 200]$, the maximum number of hidden nodes $L_{max}$ is set to 200, and sigmoid activation function is employed. Optimal hyperparameters are determined through multiple experiments using a grid search strategy. For CIL-DA, regularization parameters $C_T$ and $C_S$ are searched from set $[10^{-3}, 10^{-2}, \ldots, 10^2, 10^3]$, and $\lambda$ is searched from set $[0.1, 0.5, 1, 2, 10, 20, 50]$. The tradeoff parameters $C$ for SVM and Adapt-SVM are both searched from set $[0.1, 0.5, 1, 2, 10, 20, 50, 100]$. As for RVFL and UJD-RVFL, the radial basis function is used as their activation function and the number of enhancement layer nodes is searched from set $[50, 100, \ldots 350, 400, 500]$. Tradeoff parameters $C$ and $\lambda$ are searched from $[10^{-3}, 10^{-2}, \ldots, 10^2, 10^3]$ and $[2^{-3}, 2^{-2}, \ldots, 2^2, 2^3]$ in UJD-RVFL, respectively. Margin parameter $M$, tradeoff parameter $\gamma$ of CCSA, and tradeoff parameter $\lambda$ of CDAN are all searched from the same range of $C$ of SVM at epoch 500.

The average diagnostic accuracy and training time for all methods are reported in Table III. Note that these results are obtained by conducting the same 20 random trials for fair comparison. The detailed results unequivocally demonstrate the competitive performance of CIL-DA, surpassing all other methods in 10 out of the 12 individual domain shifts, which strongly solidifies the feasibility of the proposed domain adaptation method for bearing fault diagnosis. As a basic classier, SCN achieves an average diagnostic accuracy of approximately 84%, whereas CIL-DA consistently exhibits a higher accuracy rate and surpasses that of the SCN by around 10% in each instance, thereby underscoring the remarkable cross-domain ability of CIL-DA. It is noteworthy that CIL-DA significantly outperforms the other methods when a 3 HP motor load is used as the target domain. The reason that the restricted knowledge from the target domain is fully utilized with some auxiliary information from the source domain in CIL-DA can be believed since SCN also exhibits favorable diagnostic accuracies when utilizing the 3 HP motor load as the target domain. The methods without cross-domain ability (SCN, SVM, and RVFL) have a shorter training time than CIL-DA, which is obvious because CIL-DA has a larger amount of training samples. The modeling of CCSA and CDAN is more computationally intensive due to their utilization of deep domain adaptation techniques while CIL-DA is more lightweight and compact because of its cross-domain incremental learning. Adapt-SVM, a classical domain adaptation model based on classifiers, demonstrates exceptional modeling speed, albeit generally inferior diagnostic accuracies compared to CIL-DA across most domain shift tasks. As for UJD-RVFL, the diagnostic performance is inferior to that of CIL-DA due to the introduction of an iterative process for calculating two parameters and manually determining the number of hidden nodes.

In addition, Fig. 6 displays the confusion matrices for SCN

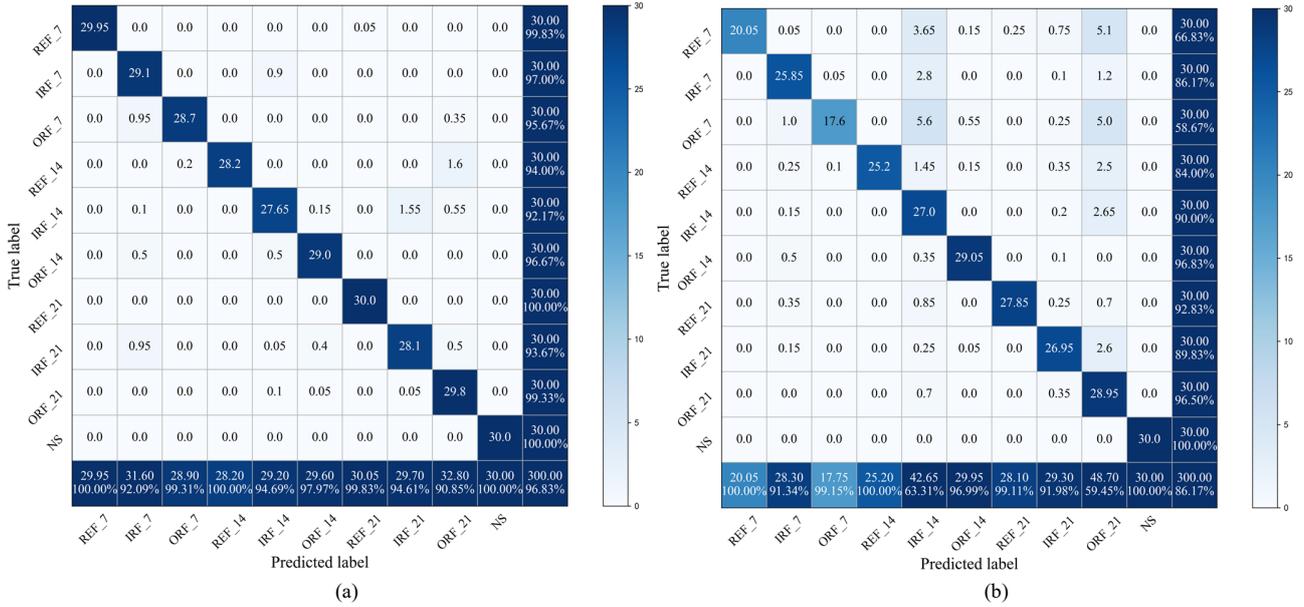

Fig. 6. Confusion matrices on case $B \to D$. The recall and precision values are written in the far-right column and far-bottom raw of picture, respectively. (a) confusion matrix of SCN. (b) confusion matrix of CIL-DA.

TABLE IV
BEARING FAULT DIAGNOSIS METHODS FOR COMPARISON

| Approach | Methodology | Abbreviation |
|---|---|---|
| Zhang et al. [3] | Deep convolutional neural Networks with wide first-layer kernels | WDCNN |
| Chen et al. [5] | Multi-scale convolutional neural network and long short-term memory | MCNN-LSTM |
| Aburakhia et al. [38] | Bayesian optimization-based random forest algorithm | Bayesian-RF |
| Li et al. [10] | Discriminative manifold random vector functional link neural network | DM-RVFLN |
| Zhang et al. [39] | A deep siamese neural network by repeating one-shot five times | Five-shot |

and CIL-DA in the $B \to D$ case. The average diagnostic accuracy of CIL-DA is 96.83%, surpassing SCN by approximately 10%. Notably, ORF with a diameter of 7 mils demonstrates the highest vulnerability to misclassification, whereas the NS health pattern is the most easily identifiable. Nonetheless, CIL-DA significantly enhances the precision and recall of ORF with a diameter of 7 mils.

### B. Comparison with Existing Fault Diagnosis Methods

This experiment compares CIL-EDA with several algorithms listed in Table IV that have performed well in bearings fault diagnosis, aiming to demonstrate the superiority of the proposed method with small fault samples. The experimental data setup for this experiment is identical to that of Experiment A, but the datasets used in Experiments A and B are not the same. It is worth mentioning that our proposed method for fault diagnosis includes a complete framework consisting of specialized feature extraction and fault classification, as depicted in Fig. 2, while the feature extraction of other methods is the same as described in their original paper. In addition, CIL-EDA addresses the issue of few-shot classification from a domain adaptation perspective via constructive incremental learning. Specifically, one of multiple operation conditions is deemed as our target domain while the remaining operation conditions serve as the source domain. This approach allows us to leverage all labeled instances from related domains effectively. Ensemble SCN (ESCN) based on Bagging is employed to prove the superior performance of CIL-EDA in bearing fault diagnosis, particularly highlighting its cross-domain capability. Note that the input of base classifiers in ESCN is also the dataset processed by the multilevel cloud feature extraction. The parameter configuration for the base classifier of ESCN is identical to that of SCN in Experiment A, and the number of base classifiers corresponds to the number of working conditions of the rolling bearing. The sigmoid function is designated as the activation function for CIL-EDA and DM-RVFLN, and the hype-parameters of base classifier of CIL-EDA are set the same as those of Experiment A. Meanwhile, the optimal number of hidden nodes of DM-RVFL is also searched from $[50,100,\dots,450,500]$, its regularization parameter $\eta$ is searched from $[10^{-3},10^{-2},\dots,10^{2},10^{3}]$, and its random weight scope is changed from $\gamma$ of CIL-DA. The training epochs and batch size of WDCNN, MCNN-LSTM and Five-shot are 5000 and 32, and their learning rate = 0.0006. For Bayesian-RF, the optimal values for $m$ and $k$ are set to 1 and according to its paper. Noteworthy, the parameter $k$ in Bayesian-RF indicates the number of decomposition levels in WPD.

The overall average results of bearing fault diagnosis, obtained by repeating each trial 20 times, are illustrated in Fig. 7. The standard deviations of diagnostic accuracies are also calculated and indicated in Fig. 7 (a). WDCNN, MCNN-LSTM, and Five-shot, as fault diagnosis methods based on deep learning, have many drawbacks, such as low classification rates and time-consuming modeling when dealing with a few training samples. MCNN-LSTM does not achieve the 90% diagnostic accuracy described in the original paper due to our insufficient training samples and its undersampling of raw signals, which may result in the loss of feature information. Additionally, the fault diagnosis methods based on deep learning exhibit diagnostic instability under multiple operation conditions. However, our proposed diagnostic method has shown superior stability and the highest accuracy compared to other methods under all operation conditions. This is mainly because the process of parallel ensemble learning strengthens learning ability by integrating multiple perspectives from individual classifiers in varying source domains, simultaneously mitigating the negative effects of improper training processes such as overtraining and inadequate training. The results of ESCN demonstrate the effectiveness of our multilevel cloud feature extraction approach. This is evidenced by the superior diagnostic performance of ESCN compared to DM-RVFLN, which extracts conventional frequency-domain features and frequency-domain features respectively, and the comparable performance to Bayesian-RF, which extracts fault-sensitive features from spectral components after WPD and also employs ensemble learning. Our CIL-EDA outperforms DM-RVFLN, which is also a single hidden layer feedforward neural network, in diagnostic accuracy by over 20% across all datasets. However, in terms of time complexity, the computational cost of CIL-EDA is slightly higher than that of other fault diagnosis methods based on shallow learner models, which can be understood because the input samples used by CIL-EDA encompass a larger amount of data from diverse source domains. In addition, the training time of CIL-EDA also exhibits less fluctuation. In general, our CIL-EDA demonstrates remarkable diagnosis performance compared to existing bearing fault diagnosis methods based on deep learner models or shallow learner models with limited samples in the target domain.

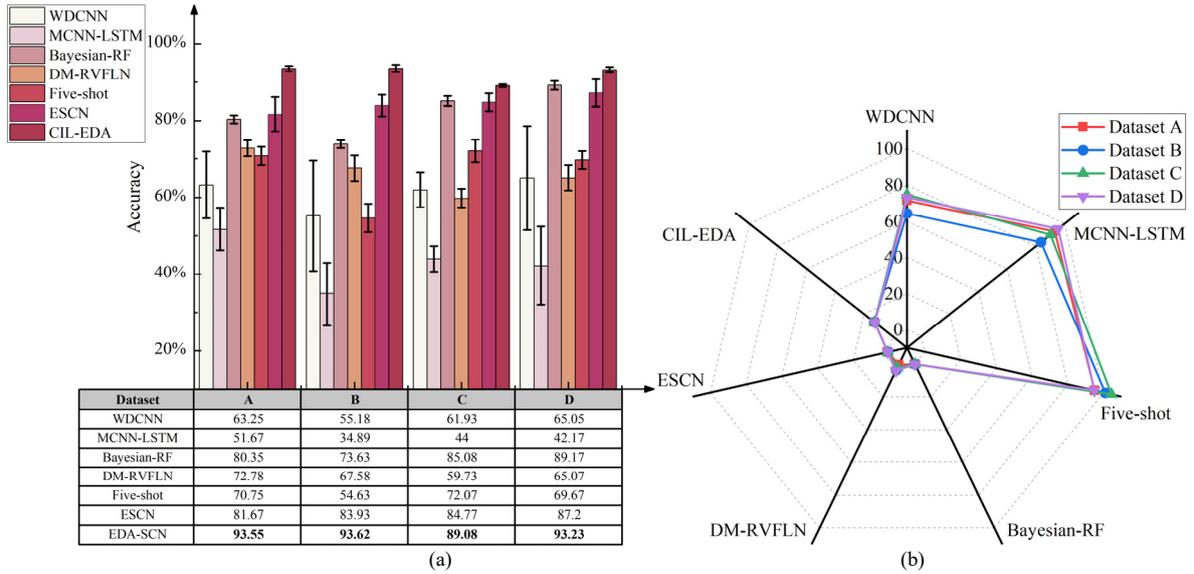

Fig. 7 Overall diagnostic performance comparisons of existing fault diagnosis methods (a) Average diagnostic accuracy. (b) Average training time.

### C. Effect of the Number of Training Samples

In this experiment, we aim to assess the effectiveness of CIL-EDA and other bearing fault diagnostic methods listed in Table IV in addressing two challenges: mechanical systems cannot be allowed to operate under faulty conditions due to their criticality and potential consequences, and most failures occur gradually over a period of months or even years, following a degradation path. Thus, we conduct a series of comparative experiments by varying the number of training samples per class belonging to the faulty state within the range of $[5, 10, 20, 30, 40]$, while maintaining a fixed number of 40 training samples in the normal state and 40 testing samples per class. It should be noted that fault samples of each quantity type include those from the preceding type. The fault diameter and health state settings of this experimental data align with those of Experiment A, with the motor load set at 2 HP. The parameters for all diagnostic methods are set in accordance with Experiment B.

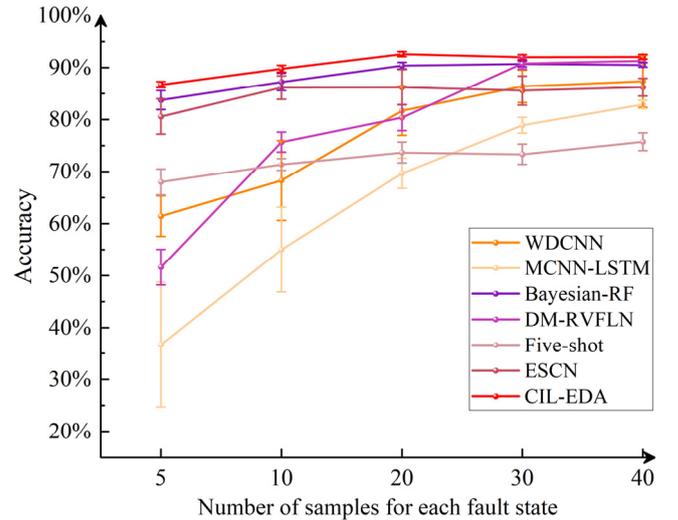

Fig. 8. Accuracy comparisons with existing fault diagnostic methods in various number of training samples

For each such sample set, we also repeat the algorithm training and testing experiment 20 times to deal with the randomness of the algorithms. Their average testing diagnostic accuracy and variance are illustrated in Fig. 8. The testing accuracy of deep learning-based diagnostic methods greatly improves with an increase in the number of training samples, indicating that the quantity of training data significantly impacts the performance of deep learner models and may lead to poor results when limited data is only available. Meanwhile, standard deviations of MCNN-LSTM and Five-shot decline with an increasing number of training samples. For these methods based on the shallow learner model, their diagnostic accuracy reaches saturation but standard deviations significantly decrease as the number of training samples per fault state increases from 20 to 40. It is clear that our proposed method consistently achieves the highest diagnostic accuracy with minimal fluctuation in various numbers of training samples. The proposed CIL-EDA performs better with an average of 6% higher in accuracy than the ESCN and demonstrates a much smaller standard deviation in various numbers of training samples, which proves that the cross-domain method based on parallel ensemble learning contributes to improving the generalization performance of the diagnostic machine. Moreover, the diagnostic accuracy of more than 80% is achieved only by CIL-EDA, Bayesian-RF, and ESCN when the number of samples for each fault state increases from 5 to 20, which declares the shallow learner model is the most appropriate approach for addressing the few-shot classification problem, with superior diagnostic accuracy and speed. These performance comparisons show that our proposed CIL-EDA enjoys a much better performance for bearing fault diagnosis when trained with limited data.

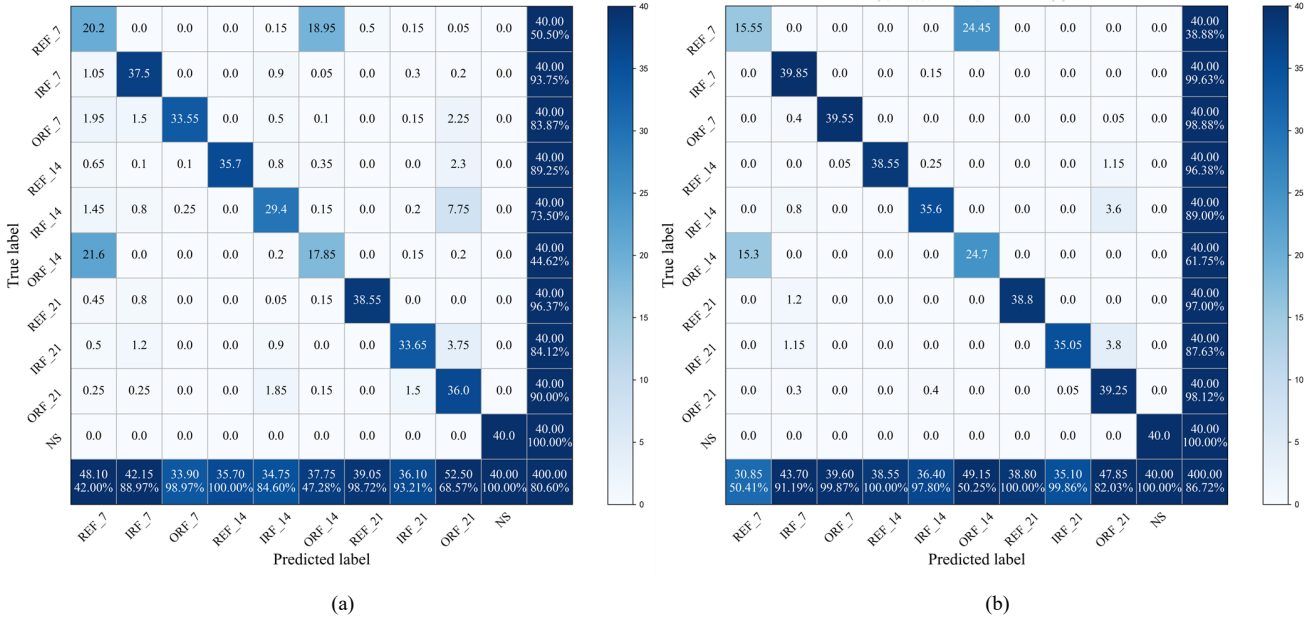

Fig. 9. Confusion matrices under 5 training samples for each fault state. (a) Confusion matrix of ESCN. (b) Confusion matrix of CIL-EDA.

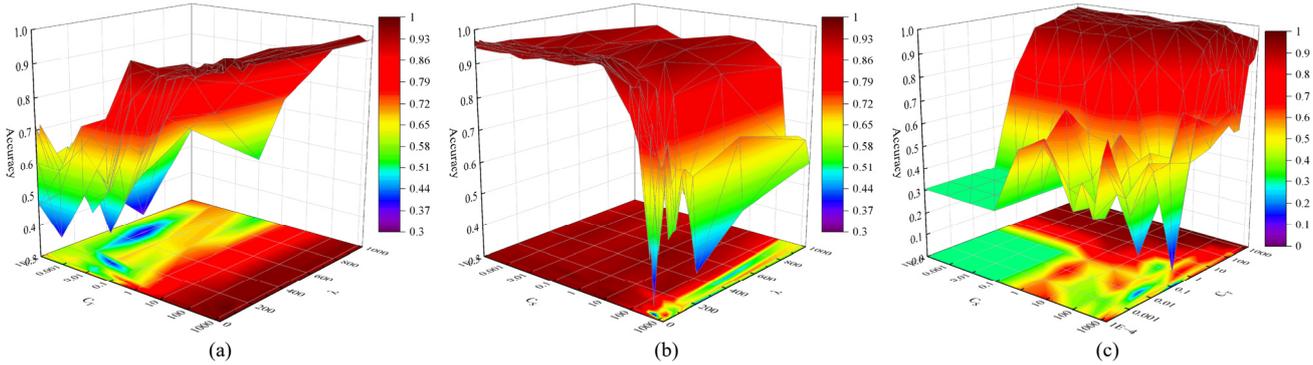

Fig. 10. Distribution of diagnostic accuracy with regularization parameters variation. (a) $C_S$ is fixed at 1. (b) $C_T$ is fixed at 100. (c) $\lambda$ is fixed at 10.

Furthermore, Fig. 9 depicts the confusion matrices of ESCN and CIL-EDA when only 5 training samples are available for each fault state. The objective is to gain a deeper understanding of the impact of cross-domain learning on fault diagnosis with limited data. CIL-EDA exhibits improved diagnostic accuracy in most fault states compared to ESCN, except for REF with 7 mils. The key distinction between the two methods lies in CIL-EDA's comprehensive utilization of existing data from multiple working conditions and its implementation of domain matching, which facilitates effective transfer learning across diverse domains. In contrast, ESCN solely relies on ensemble learning based on Bagging and lacks the capability for cross-domain learning. Once again, this experiment demonstrates the significant role of CIL-EDA in bearing fault diagnosis with limited sample sizes.

*D. Parameter sensitivity*

In this section, empirical parameter sensitivity analysis of three regularization parameters $C_S$, $C_T$ and $\lambda$ involved in CIL-DA is conducted. Due to space limitation, one domain shift B $\rightarrow$ A is selected for sensitivity analysis. Only two parameters are allowed to change with the last parameter fixed each time, while the fixed parameters are chosen as $C_S=1$, $C_T=100$ and $\lambda=10$. The results are shown as Fig. 12 and a concise analysis is provided here. For $C_S$ and $C_T$, they are used for balancing the contributions of source and target domain. The model gains more knowledge from the target domain when $C_S$ is smaller than $C_T$. And $\lambda$ is used to control the differences of cross-domain matching and adjusts the domain adaptive ability of the model. Since the number of samples in source domain is the same as that of samples in target domain in this experiment, the reasonable $C_T$ should be larger than 1 and $C_S$ for higher accuracy in our target working condition, as shown in Fig. 12(c). $\lambda$ cannot be too large and its appropriate range should be [1,100], which is consistent with the experimental results shown in Fig. 12(a) and (b).

## VI. CONCLUSION

In this paper, CIL-DA is proposed to introduce a perspective on constructive incremental domain adaptation, followed by the development of CIL-EDA to address the fault diagnosis issue for rolling bearings under multiple working conditions with a limited number of target samples. CIL-DA is capable of maintaining the inter-class relationship of the source domain by unilaterally aligning the source domain to the target domain, instead of aligning the discrepancies for the source and target domain by mapping both of their feature spaces into an unknown intermediate space. Besides, CIL-DA simultaneously aligns both marginal and conditional distribution discrepancies during the incremental learning of hidden nodes, rather than incremental learning of input data. Building upon CIL-DA, CIL-EDA applies ensemble learning to leverage the available limited data from multiple source domains, thereby enhancing the generalization performance of the base classifier in the target domain.

To verify the effectiveness and feasibility of our proposed approaches, extensive experiments have been performed on the CWRU Datasets, including transfer tests for CIL-DA, fault diagnostic tests for CIL-EDA, effect tests with various numbers of training samples, and parameter sensitivity tests for CIL-DA. The experimental results demonstrate the effectiveness of the CIL-DA cross-domain approach by comparing its performance against baselines established using both deep learner models and shallow learner models. Furthermore, the overall diagnostic results strongly support the capability of CIL-EDA to serve as a potential fault diagnostic technique for rolling bearings in various industrial applications, regardless of the availability of limited or sufficient data.

The integration of data from multiple perspectives can offer a holistic comprehension of a target object. In the future, multi-view learning combined with transfer learning is worth of further study in incremental learning model.